\documentclass[11pt,a4paper,dvipsnames]{article}
\usepackage[hyperref]{naaclhlt2019}
\usepackage{times}
\usepackage{latexsym}

\usepackage{url}

\aclfinalcopy 

\title{Instructions for NAACL-HLT 2019 Proceedings}

\author{Pouya Pezeshkpour \\
  University of California \\
  Irvine, CA \\
  \href{mailto:pezeshkp@uci.edu}{\tt pezeshkp@uci.edu} \\\And
  Yifan Tian \\
  University of California \\
  Irvine, CA \\
  \href{mailto:yifant@uci.edu}{\tt yifant@uci.edu} \\\And
  Sameer Singh \\
  University of California \\
  Irvine, CA \\
  \href{mailto:sameer@uci.edu}{\tt sameer@uci.edu} \\
  }

\date{}
\newcommand{\conference}{NAACL}
\usepackage{graphicx} 
\usepackage{ifthen}
\ifthenelse{ \equal{\conference}{AKBC} }{
    \usepackage[dvipsnames]{xcolor}
    \usepackage{hyperref}       
}{
}
\usepackage{epsfig} 
\usepackage[utf8]{inputenc} 
\usepackage[T1]{fontenc}    
\usepackage{url}            
\usepackage{booktabs}       
\usepackage{amsfonts}       
\usepackage{nicefrac}       
\usepackage{microtype}      
\usepackage[inline]{enumitem}

\usepackage{algorithm}
\usepackage{algorithmic}
\usepackage{float}
\usepackage{tikz}

\usepackage{latexsym}
\usepackage{amsmath}
\usepackage{amsthm}
\usepackage{amssymb}
\usepackage{mathtools}
\usepackage{multirow}
\usepackage{bm}
\usepackage{soul}
\usepackage{lscape}
\usepackage{graphicx,epstopdf}
\usepackage{psfrag}
\usepackage{rotating}
\usepackage{outlines}
\usepackage[textsize=small]{todonotes}
\usepackage{subcaption}
\usepackage{textcomp}
\usepackage{wrapfig}
\usepackage{xspace}
\usetikzlibrary{shapes,arrows,patterns}
\usetikzlibrary{positioning}
\usetikzlibrary{calc}

\newcommand{\AALP}{CRIAGE\xspace}
\newcommand{\AMLPFT}{\AALP-FT\xspace}
\newcommand{\AALPA}{\AALP-Add\xspace}
\newcommand{\AALPR}{\AALP-Remove\xspace}
\newcommand{\AALPBE}{\AALP-Best\xspace}

\newcommand{\triple}[1]{\langle #1\rangle}
\newcommand{\vs}{\mathbf{e}_s}
\newcommand{\vr}{\mathbf{e}_r}
\newcommand{\vo}{\mathbf{e}_o}
\newcommand{\z}{\mathbf{z}}
\newcommand{\x}{\mathbf{x}}
\newcommand{\vsn}{\overline{\mathbf{e}_s}}
\newcommand{\vrn}{\overline{\mathbf{e}_r}}
\newcommand{\von}{\overline{\mathbf{e}_o}}
\newcommand{\f}{\mathbf{f}}
\newcommand{\g}{\mathbf{g}}
\newcommand{\loss}{\mathcal{L}}

\newcommand{\relLabel}[1]{\texttt{#1}}

\newcommand{\sameer}[1]{\textcolor{red}{\textbf{s:} #1}}

\newcommand{\entity}[1]{\texttt{#1}}

\newcommand{\para}[1]{\vspace{1.5mm}\noindent\textbf{#1} }

\newcommand{\R}{\ensuremath{\mathcal{R}}}
\newcommand{\Real}{\mathbb{R}}

\newcommand{\cut}[1]{}

\begin{document}

\title{Investigating Robustness and Interpretability of Link Prediction\\ via Adversarial Modifications}

\maketitle
\begin{abstract}
Representing entities and relations in an embedding space is a well-studied approach for machine learning on relational data. Existing approaches, however, primarily focus on improving accuracy and overlook other aspects such as robustness and interpretability. In this paper, we propose adversarial modifications for link prediction models: identifying the fact to add into or remove from the knowledge graph that changes the prediction for a target fact after the model is retrained. Using these single modifications of the graph, we identify the most influential fact for a predicted link and evaluate the sensitivity of the model to the addition of fake facts. We introduce an efficient approach to estimate the effect of such modifications by approximating the change in the embeddings when the knowledge graph changes. To avoid the combinatorial search over all possible facts, we train a network to \emph{decode} embeddings to their corresponding graph components, allowing the use of gradient-based optimization to identify the adversarial modification. We use these techniques to evaluate the robustness of link prediction models (by measuring sensitivity to additional facts), study interpretability through the facts most responsible for predictions (by identifying the most influential neighbors), and detect incorrect facts in the knowledge base. 

\end{abstract}


 \begin{figure*}[tb] 
  \begin{center}
    \begin{subfigure}[b]{0.31\textwidth}
        \centering
        \begin{tikzpicture}[scale=0.6, every node/.style={scale=0.6}]
            \node[draw,circle,thick,inner sep=0,minimum size=2.1cm] (o_fm) at (0,0) {\begin{tabular}{c} Ferdinand \\ Maria \end{tabular}};
            \node[draw,circle,thick,inner sep=0,minimum size=2.1cm] (o_ph) at (4,0) {\begin{tabular}{c} Princess\\ Henriette\end{tabular}};
            \node[draw,circle,thick,inner sep=0,minimum size=2.1cm] (o_vb) at (0,-4) {\begin{tabular}{c} Violante\\ Bavaria\end{tabular}};
    
            \draw[->,thick] (o_fm) -- (o_ph) node[pos=0.5, sloped,align=center, above]{isMarried};
            \draw[->,thick] (o_fm) -- (o_vb) node[pos=0.5, sloped,align=center,rotate=180, above]{hasChild};
            \draw[->,thick,blue,densely dashed] (o_ph) -- (o_vb) node[pos=0.5, sloped,align=center, above] {hasChild};
            
            \node (a) at (3.5,-3.5) {\begin{tabular}{c} target prediction\\ $\langle s,r,o\rangle$\end{tabular} };
            \draw[->] (a) -- (2.1,-2.1);
        \end{tikzpicture}
        \caption{KG, with the target prediction}
        \label{fig:AALP:orig}
    \end{subfigure}
    \begin{subfigure}[b]{0.31\textwidth}
        \centering
        \begin{tikzpicture}[scale=0.6, every node/.style={scale=0.6}]
            \node[draw,circle,thick,inner sep=0,minimum size=2.1cm] (r_fm) at (12,-5) {\begin{tabular}{c} Ferdinand \\ Maria \end{tabular}};
            \node[draw,circle,thick,inner sep=0,minimum size=2.1cm] (r_ph) at (16,-5) {\begin{tabular}{c} Princess\\ Henriette\end{tabular}};
            \node[draw,circle,thick,inner sep=0,minimum size=2.1cm] (r_vb) at (12,-9) {\begin{tabular}{c} Violante\\ Bavaria\end{tabular}};
            \node[draw,circle,thick,inner sep=0,minimum size=2.1cm] (r_ny) at (16,-9) {\begin{tabular}{c} A.S.D.\\Astrea\end{tabular}};
            
            \draw[->,thick] (r_fm) -- (r_ph) node[pos=0.5, sloped,align=center, above]{isMarried};
            \draw[->,thick,red!50!black] (r_fm) -- (r_vb) node[pos=0.5, sloped,align=center, above]{hasChild} node[pos=0.5, sloped,align=center, below]{$\langle s',r',o\rangle$\\\textbf{removed}};
            \draw[->,thick,blue,densely dashed] (r_ph) -- (r_ny) node[pos=0.5, sloped,align=center, above] {hasChild};
        \end{tikzpicture}
        \caption{After \textcolor{red!50!black}{removing} a fact}
        \label{fig:AALP:remove}
    \end{subfigure}
    \begin{subfigure}[b]{0.35\textwidth}
        \centering
        \begin{tikzpicture}[scale=0.6, every node/.style={scale=0.6}]
            \node[draw,circle,thick,inner sep=0,minimum size=2.1cm] (a_fm) at (12,5) {\begin{tabular}{c} Ferdinand \\ Maria \end{tabular}};
            \node[draw,circle,thick,inner sep=0,minimum size=2.1cm] (a_ph) at (16,5) {\begin{tabular}{c} Princess\\ Henriette\end{tabular}};
            \node[draw,circle,thick,inner sep=0,minimum size=2.1cm] (a_vb) at (12,1) {\begin{tabular}{c} Violante\\ Bavaria\end{tabular}};
            \node[draw,circle,thick,inner sep=0,minimum size=2.1cm] (a_ny) at (18.5,2.5) {\begin{tabular}{c} New\\ York\end{tabular}};
            \node[draw,green!50!black,circle,thick,inner sep=0,minimum size=2.1cm] (a_aj) at (16,1) {\begin{tabular}{c} Al Jazira\\ Club\end{tabular}};
            
            \draw[->,thick] (a_fm) -- (a_ph) node[pos=0.5, sloped,align=center, above]{isMarried};
            \draw[->,thick] (a_fm) -- (a_vb) node[pos=0.5, sloped,align=center, above]{hasChild};
            \draw[->,thick,green!50!black] (a_aj) -- (a_vb) node[pos=0.5, sloped,align=center, above]{playsFor} node[pos=0.5, sloped,align=center, below]{$\langle s',r',o\rangle$\\\textbf{added}};;
            \draw[->,thick,blue,densely dashed] (a_ph) -- (a_ny) node[pos=0.5, sloped,align=center, above] {hasChild};
        \end{tikzpicture}
        \caption{After \textcolor{green!50!black}{adding} a fact}
        \label{fig:AALP:add}
    \end{subfigure}
  \end{center}
  \caption{\textbf{Completion Robustness and Interpretability via Adversarial Graph Edits (\AALP):} Change in the graph structure that changes the prediction of the \emph{retrained} model, where (a) is the original sub-graph of the KG, (b) removes a neighboring link of the target, resulting in a change in the prediction, and (c) shows the effect of adding an attack triple on the target. These modifications were identified by our proposed approach.} 
  \label{fig:AALP}
\end{figure*}
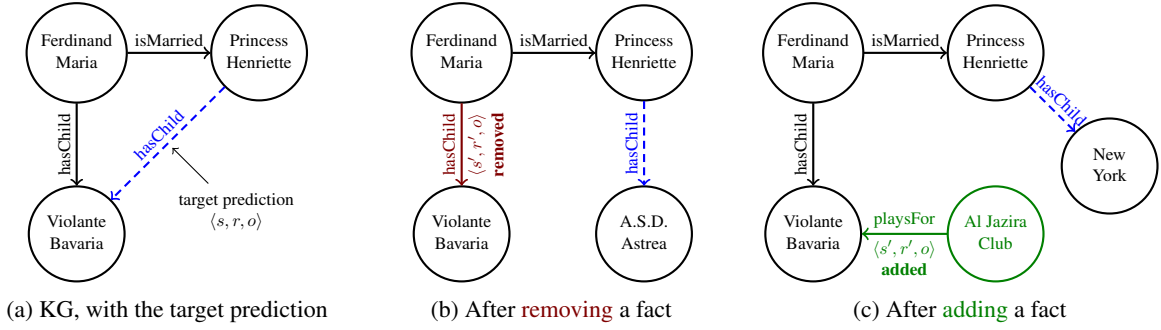

\section{Introduction}
Knowledge graphs (KG) play a critical role in many real-world applications such as search, structured data management, recommendations, and question answering. 
Since KGs often suffer from incompleteness and noise in their facts (links), a number of recent techniques have proposed models that \emph{embed}
each entity and relation into a vector space, and use these embeddings to predict facts. 
These dense representation models for link prediction include tensor factorization~\citep{nickel2011three, socher2013reasoning, yang2014embedding}, algebraic operations~\citep{bordes2011learning,bordes2013translating,dasgupta2018hyte}, multiple embeddings~\citep{wang2014knowledge,lin2015learning,ji2015knowledge,zhang2018knowledge}, and complex neural models~\citep{dettmers2017convolutional,nguyen2018novel}. 
However, there are only a few studies~\citep{kadlec2017knowledge,sharma2018towards} that investigate the quality of the different KG models. 
There is a need to go beyond just the accuracy on link prediction, and instead focus on whether these representations are robust and stable, and what facts they make use of for their predictions. 

In this paper, our goal is to design approaches that minimally change the graph structure such that the prediction of a target fact changes the most after \emph{the embeddings are relearned}, which we collectively call \emph{Completion Robustness and Interpretability via Adversarial Graph Edits} (\AALP). 
First, we consider perturbations that \textcolor{red!50!black}{remove a neighboring link} for the target fact, thus identifying the \emph{most influential} related fact, providing an explanation for the model's prediction.
As an example, consider the excerpt from a KG in Figure~\ref{fig:AALP:orig} with two observed facts, and a target \emph{predicted fact} that \entity{Princes Henriette} is the parent of \entity{Violante Bavaria}. Our proposed graph perturbation, shown in Figure~\ref{fig:AALP:remove}, identifies the existing fact that \entity{Ferdinal Maria} is the father of \entity{Violante Bavaria} as the one when removed and model retrained, will change the prediction of \entity{Princes Henriette}'s child.
We also study attacks that \textcolor{green!50!black}{add a new, fake fact} into the KG to evaluate the robustness and sensitivity of link prediction models to small additions to the graph. 
An example attack for the original graph in Figure~\ref{fig:AALP:orig}, 
is depicted in Figure \ref{fig:AALP:add}. 
Such perturbations to the the training data are from a family of adversarial modifications that have been applied to other machine learning tasks, known as \emph{poisoning}~\citep{biggio2012poisoning,corona2013adversarial,biggio2014security,zugner2018adversarial}. 

Since the setting is quite different from traditional adversarial attacks, search for link prediction adversaries brings up unique challenges.
To find these minimal changes for a target link, we need to identify the fact that, when added into or removed from the graph, will have the biggest impact on the predicted score of the target fact. 
Unfortunately, computing this change in the score is expensive since it involves retraining the model to recompute the embeddings. 
We propose an efficient estimate of this score change by approximating the change in the embeddings using Taylor expansion.
The other challenge in identifying adversarial modifications for link prediction, especially when considering addition of fake facts, is the combinatorial search space over possible facts, which is intractable to enumerate. 
We introduce an \emph{inverter} of the original embedding model, to \emph{decode} the embeddings to their corresponding graph components, making the search of facts tractable by performing efficient gradient-based continuous optimization. 

We evaluate our proposed methods through following experiments. First, 
on relatively small KGs, we show that our approximations are accurate compared to the true change in the score. 
Second, we show that our additive attacks can effectively reduce the performance of state of the art models~\citep{yang2014embedding, dettmers2017convolutional} up to $27.3\%$ and $50.7\%$ in Hits@1 for two large KGs: WN18 and YAGO3-10.
We also explore the utility of adversarial modifications in explaining the model predictions 
by presenting rule-like descriptions of the most influential neighbors. 
Finally, we use adversaries to detect errors in the KG, 
obtaining up to $55\%$ accuracy in detecting errors.

\section{Background and Notation}
In this section, we briefly introduce some notations, and existing relational embedding approaches that model knowledge graph completion using dense vectors.
%
In KGs, facts are represented using triples of subject, relation, and object, $\langle s, r, o\rangle$, where $s,o\in\xi$, the set of entities, and $r\in \R$, the set of relations. 
To model the KG, a scoring function $\psi:\xi\times \R\times\xi\rightarrow\Real$ is learned to evaluate whether any given fact is true. 
%
In this work, we focus on \emph{multiplicative} models of link prediction\footnote{As opposed to \emph{additive} models, such as TransE~\citep{bordes13:translating}, as categorized in \citet{sharma2018towards}.}, specifically DistMult~\citep{yang2014embedding} because of its simplicity and popularity, and ConvE~\citep{dettmers2017convolutional} because of its high accuracy. 
We can represent the scoring function of such methods as  $\psi(s,r,o) = \f(\vs, \vr) \cdot \vo$, where $\vs,\vr,\vo\in \Real^d$ are embeddings of the subject, relation, and object respectively. 
In DistMult, $\f(\vs, \vr) = \vs \odot \vr$, where $\odot$ is element-wise multiplication operator. 
Similarly, in ConvE, $\f(\vs, \vr)$ is computed by a convolution on the concatenation of $\vs$ and $\vr$.    
We use the same setup as \citet{dettmers2017convolutional} for training, i.e., incorporate binary cross-entropy loss over the triple scores. 
In particular, for subject-relation pairs $(s,r)$ in the training data $G$, we use binary $y^{s,r}_o$ to represent negative and positive facts. 
Using the model's probability of truth as $\sigma(\psi(s,r,o))$ for $\langle s,r,o\rangle$, the loss is defined as:
\begin{align}
    \loss(G) &= \sum_{(s,r)}\sum_{o} y^{s,r}_o\log(\sigma(\psi(s,r,o)))\nonumber\\
    &+ (1-y^{s,r}_o)\log(1 - \sigma(\psi(s,r,o))).
\end{align}
Gradient descent is used to learn the embeddings $\vs,\vr,\vo$, and the parameters of $\f$, if any.

\section{Completion Robustness and Interpretability via Adversarial Graph Edits (\AALP)}
For adversarial modifications on KGs, we first define the space of possible modifications. 
For a target triple $\langle s, r, o\rangle$, we constrain the possible triples that we can remove (or inject) to be in the form of $\langle s', r', o\rangle$ i.e $s'$ and $r'$ may be different from the target, but the object is not.
We analyze other forms of modifications such as $\langle s, r', o'\rangle$ and $\langle s, r', o\rangle$ in appendices \ref{app:srpop} and \ref{app:srpo}, and leave empirical evaluation of these modifications for future work. 


\subsection{Removing a fact (\AALPR)}
\label{sec:aalp:remove}

For explaining a target prediction, we are interested in identifying the observed fact that has the most influence (according to the model) on the prediction.
We define \emph{influence} of an observed fact on the prediction as the change in the prediction score if the observed fact was not present when the embeddings were learned. Previous work have used this concept of influence similarly for several different tasks~\citep{kononenko2010efficient,koh2017understanding}. 
%
Formally, for the target triple $\triple{s,r,o}$ and observed graph $G$, we want to identify a neighboring triple $\triple{s',r',o}\in G$ such that the score $\psi(s,r,o)$ when trained on $G$ and the score $\overline{\psi}(s,r,o)$ when trained on $G-\{\triple{s',r',o}\}$ are maximally different, i.e.
\begin{align}
   \operatorname*{argmax}_{(s', r')\in \text{Nei}(o)} \Delta_{(s',r')}(s,r,o)
\end{align}
where $\Delta_{(s',r')}(s,r,o)=\psi(s, r, o)-\overline{\psi}(s,r,o)$, and
$\text{Nei}(o)=\{(s',r')|\langle s',r',o \rangle \in G \}$. 

\subsection{Adding a new fact (\AALPA)}
\label{sec:aalp:add}

We are also interested in investigating the robustness of models, i.e., how sensitive are the predictions to small additions to the knowledge graph. Specifically, for a target prediction $\triple{s,r,o}$, we are interested in identifying a single fake fact $\triple{s',r',o}$ that, when added to the knowledge graph $G$, changes the prediction score $\psi(s,r,o)$ the most.
Using $\overline{\psi}(s,r,o)$ as the score after training on $G\cup\{\triple{s',r',o}\}$, we define the adversary as:
\begin{align}
   \operatorname*{argmax}_{(s', r')} \Delta_{(s',r')}(s,r,o)  
\end{align}
where $\Delta_{(s',r')}(s,r,o)=\psi(s, r, o)-\overline{\psi}(s,r,o)$. The search here is over any possible $s'\in\xi$, which is often in the millions for most real-world KGs, and $r'\in\R$. 
We also identify adversaries that \emph{increase} the prediction score for specific false triple, i.e., for a target fake fact $\triple{s,r,o}$, 
the adversary is
$   \operatorname*{argmax}_{(s', r')} - \Delta_{(s',r')}(s,r,o) 
$,
where $\Delta_{(s',r')}(s,r,o)$ is defined as before.

\subsection{Challenges}
There are a number of crucial challenges when conducting such adversarial attack on KGs. 
First, evaluating the effect of changing the KG on the score of the target fact ($\overline{\psi}(s,r,o)$) is expensive since we need to update the embeddings by retraining the model on the new graph; a very time-consuming process that is at least linear in the size of $G$. 
Second, since there are many candidate facts that can be added to the knowledge graph, identifying the most promising adversary through search-based methods is also expensive. 
Specifically, the search size for unobserved facts is $|\xi| \times |\R|$, which, for example in YAGO3-10 KG, can be as many as $4.5 M$ possible facts for a single target prediction. 
\section{Efficiently Identifying the Modification} 
\label{app_at}

In this section, we propose algorithms to address mentioned challenges by (1) approximating the effect of changing the graph on a target prediction, and (2) using continuous optimization for the discrete search over potential modifications. 

\subsection{First-order Approximation of Influence}

We first study the addition of a fact to the graph, and then extend it to cover removal as well. 
To capture the effect of an adversarial modification on the score of a target triple, we need to study the effect of the change on the vector representations of the target triple. 
We use $\vs$, $\vr$, and $\vo$ to denote the embeddings of $s,r,o$ at the solution of $\operatorname*{argmin} \loss(G)$, and when considering the adversarial triple $\langle s', r', o \rangle $, we use $\vsn$, $\vrn$, and $\von$ for the new embeddings of $s,r,o$, respectively. 
Thus $\vsn,\vrn,\von$ is a solution to $\operatorname*{argmin} \loss(G\cup\{\langle s', r', o \rangle\})$, which can also be written as $\operatorname*{argmin} \loss(G)+\loss(\langle s', r', o \rangle)$. 
Similarly, ${\f}(\vs, \vr)$ changes to $\f{(\vsn, \vrn)}$ after retraining. 

Since we only consider adversaries in the form of $\langle s', r', o \rangle $, we only consider the effect of the attack on $\vo$ and neglect its effect on $\vs$ and $\vr$. 
This assumption is reasonable since the adversary is connected with $o$ and directly affects its embedding when added, but it will only have a secondary, negligible effect on $\vs$ and $\vr$, in comparison to its effect on $\vo$. 
Further, calculating the effect of the attack on $\vs$ and $\vr$ requires a third order derivative of the loss, which is not practical~($O(n^3)$ in the number of parameters). 
In other words, we assume that $\vsn \simeq \vs$ and $\vrn \simeq \vr$. 
As a result, to calculate the effect of the attack, 
$\overline{\psi}{(s,r,o)}-\psi(s, r, o)$, we need to compute $\von-\vo$, followed by: 
\begin{align}
\overline{\psi}{(s,r,o)}-\psi(s, r, o)= \z_{s, r}(\von-\vo)
\end{align}
where $\z_{s, r} = \f(\vs,\vr)$. 
We now derive an efficient computation for $\von-\vo$. 
First, the derivative of the loss $\loss(\overline{G})= \loss(G)+\loss(\langle s', r', o \rangle)$ over $\vo$ is:
 \begin{align}
    \nabla_{e_o} \loss(\overline{G}) = \nabla_{e_o} \loss(G) - (1-\varphi) \z_{s', r'} 
 \end{align}
 where $\z_{s', r'} = \f(\vs',\vr')$, and $\varphi = \sigma(\psi(s',r',o))$. 
At convergence, after retraining, we expect $\nabla_{e_o} \loss(\overline{G})=0$.
We perform first order Taylor approximation of $\nabla_{e_o} \loss(\overline{G})$ to get:
\begin{align}
    0 \simeq &- (1-\varphi)\z_{s',r'}^\intercal+&\nonumber\\
    &(H_o+\varphi(1-\varphi)\z_{s',r'}^{\intercal} \z_{s',r'})(\von-\vo)&
\end{align}
where $H_o$ is the $d\times d$ Hessian matrix for $o$, i.e., second order derivative of the loss w.r.t. $\vo$, computed sparsely.
Solving for $\von-\vo$ gives us, $\von-\vo=$:
\begin{align}
    &(1-\varphi) (H_o + \varphi (1-\varphi) \z_{s',r'}^\intercal\z_{s',r'})^{-1} \z_{s',r'}^\intercal. \nonumber
\end{align}
In practice, $H_o$ is positive definite, making $H_o + \varphi (1-\varphi) \z_{s',r'}^\intercal\z_{s',r'}$ positive definite as well, and invertible. Then, we compute the score change as: 
\begin{align}
       &\overline{\psi}{(s,r,o)}-\psi(s, r, o)= \z_{s,r} (\von-\vo) &\label{eq:approx:add}\\
       &= \z_{s,r} ((1-\varphi) (H_o + \varphi (1-\varphi) \z_{s',r'}^\intercal \z_{s',r'})^{-1} \z_{s',r'}^\intercal ).&\nonumber
\end{align}
Calculating this expression is efficient since $H_o$ is a $d\times d$ matrix ($d$ is the embedding dimension), and $\z_{s,r},\z_{s',r'}\in \Real^d$.
%
Similarly, we estimate the score change of $\langle s, r, o \rangle$ after \emph{removing} $\langle s', r', o \rangle$ as: 
\[
    -  \z_{s,r} ((1-\varphi) (H_o + \varphi (1-\varphi) \z_{s',r'}^\intercal \z_{s',r'})^{-1} \z_{s',r'}^\intercal ).\label{eq:approx:remove}
\]

 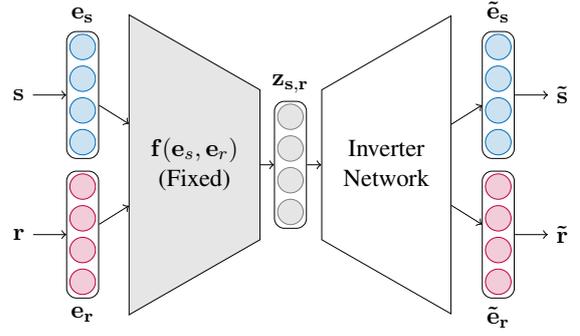
\begin{figure}[tb] 
  \begin{center}
    \resizebox{\columnwidth}{!}{
    \begin{tikzpicture}[scale=0.9, every node/.style={scale=0.9}]
\tikzset{mysymbol/.pic={
    \node[rectangle, rounded corners, minimum width=0.5cm, minimum height=2cm,text centered, draw=black] (-sum) {};
    \filldraw[fill opacity=0.2] ([yshift=-0.25cm]-sum.north) circle (0.2cm);
    \filldraw[fill opacity=0.2] ([yshift=-0.75cm]-sum.north) circle (0.2cm);
    \filldraw[fill opacity=0.2] ([yshift=-1.25cm]-sum.north) circle (0.2cm);
    \filldraw[fill opacity=0.2] ([yshift=-1.75cm]-sum.north) circle (0.2cm);
}}

\node (s) at (-4,1.1) {$\mathbf{s}$};
\path pic[NavyBlue] (vects) at ([xshift=1cm]s) {mysymbol};
\node[above=0mm of vects-sum] (tvects) {$\mathbf{e_s}$};
\draw[->] (s) -- (vects-sum);

\node (r) at (-4,-1.1) {$\mathbf{r}$};
\path pic[purple] (vectr) at ([xshift=1cm]r) {mysymbol};
\node[below=0mm of vectr-sum] (tvectr) {$\mathbf{e_r}$};
\draw[->] (r) -- (vectr-sum);

\node[draw,inner ysep=20pt,fill=black!10!white,trapezium,rotate=0,shape border rotate=270] (enc) at (-1.25,0) {\begin{tabular}{c} $\f(\vs,\vr)$\\ (Fixed) \end{tabular}};
\draw[->] (vectr-sum) -- (enc);
\draw[->] (vects-sum) -- (enc);

\path pic[gray] (vectz) at ([xshift=1.5cm]enc) {mysymbol};
\draw[->] (enc) -- (vectz-sum);
\node[above=0mm of vectz-sum] (tvectz) {$\mathbf{z_{s,r}}$};

\node[draw,inner ysep=20pt,trapezium,rotate=0,shape border rotate=90] (dec) at (1.75,0) {\begin{tabular}{c} Inverter \\ Network \end{tabular}};
\draw[->] (vectz-sum) -- (dec);

\node (sp) at (4.5,1.1) {$\mathbf{\tilde{s}}$};
\path pic[NavyBlue] (vectsp) at ([xshift=-1cm]sp) {mysymbol};
\node[above=0mm of vectsp-sum] (tvectsp) {$\mathbf{\tilde{e}_s}$};
\draw[<-] (sp) -- (vectsp-sum);

\node (rp) at (4.5,-1.1) {$\mathbf{\tilde{r}}$};
\path pic[purple] (vectrp) at ([xshift=-1cm]rp) {mysymbol};
\node[below=0mm of vectrp-sum] (tvectrp) {$\mathbf{\tilde{e}_r}$};
\draw[<-] (rp) -- (vectrp-sum);

\draw[<-] (vectrp-sum) -- (dec);
\draw[<-] (vectsp-sum) -- (dec);




\end{tikzpicture}
    }
  \end{center}
  \caption{\textbf{Inverter Network} The architecture of our inverter function 
  that translate $\z_{s, r}$ to its respective $(\tilde s,\tilde r)$. The encoder component is fixed to be the encoder network of DistMult and ConvE respectively. }
  \label{fig:auto}
\end{figure}

\subsection{Continuous Optimization for Search}
Using the approximations provided in the previous section, Eq.~\eqref{eq:approx:add} and \eqref{eq:approx:remove}, we can use brute force enumeration to find the adversary $\langle s', r', o \rangle$.
This approach is feasible when removing an observed triple since the search space of such modifications is usually small; it is the number of observed facts that share the object with the target.
On the other hand, finding the most influential unobserved fact to add requires search over a much larger space of all possible unobserved facts (that share the object). 
Instead, we identify the most influential unobserved fact $\langle s', r', o \rangle$ by using a gradient-based algorithm on vector $\z_{s',r'}$ in the embedding space (reminder,  $\z_{s',r'}=\f(\vs',\vr')$), solving the following continuous optimization problem in $\Real^d$: 
\begin{align}
   \operatorname*{argmax}_{\z_{s', r'}} \Delta_{(s',r')}(s,r,o).
\end{align}
After identifying the optimal $\z_{s', r'}$, we still need to generate the pair $(s',r')$.
We design a network, shown in Figure~\ref{fig:auto}, that maps the vector $\z_{s',r'}$ to the entity-relation space, i.e., translating it into $(s',r')$. 
In particular, we train an auto-encoder where the encoder is fixed to receive the $s$ and $r$ as one-hot inputs, and calculates $\z_{s, r}$ in the same way as the DistMult and ConvE encoders respectively (using trained embeddings).
The decoder is trained to take $\z_{s,r}$ as input and produce $s$ and $r$, essentially inverting $\f$ and the embedding layers. 
As our decoder, for DistMult, we pass $\z_{s, r}$ through a linear layer and then use two other linear layers for the subject and the relation separately, providing one-hot vectors as $\tilde{s}$ and $\tilde{r}$. 
For ConvE, we pass $\z_{s, r}$ through a deconvolutional layer, and then use the same architecture as the DistMult decoder. 
Although we could use maximum inner-product search \citep{shrivastava2014asymmetric} for DistMult instead of our defined inverter function, we are looking for a general approach that works across multiple models.

We evaluate the performance of our inverter networks (one for each model/dataset) on correctly recovering the pairs of subject and relation from the test set of our benchmarks, given the $\z_{s,r}$. The accuracy of recovered pairs (and of each argument) is given in Table~\ref{tab:inverter-acc}. 
As shown, our networks achieve a very high accuracy, demonstrating their ability to invert vectors $\z_{s,r}$ to $\{s,r\}$ pairs.

\begin{table}[tb]
    \centering
    \resizebox{0.48\textwidth}{!}{
    \small
    \begin{tabular}{lccccc}
    \toprule
        \multirow{2}{*}{ }&\multicolumn{2}{c}{\bf             WordNet}&\multicolumn{2}{c}{\bf YAGO}\\
        \cmidrule(lr){2-3}
        \cmidrule(lr){4-5}
        &DistMult&ConvE&DistMult&ConvE\\
    \midrule
        Recover $s$&93.4&96.1&97.2&98.1\\
        Recover $r$ & 91.3&95.3&99.0&99.6\\
        Recover $\{s,r\}$ &89.5&94.2&96.4&98.0\\
    \bottomrule
    \end{tabular}
     }
    \caption{\textbf{Inverter Functions Accuracy}, we calculate the accuracy of our inverter networks in correctly recovering the pairs of subject and relation from the test set of our benchmarks.}
    \label{tab:inverter-acc}
\end{table}

\begin{table}[tb]
  \centering
        \small
        \begin{tabular}{lrrrr}
        \toprule
        & \bf \# Rels & \bf \#Entities & \bf \# Train & \bf \#Test \\
        \midrule
        Nations & 56&14&1592&200 \\
        Kinship & 26 & 104&4,006& 155\\
        WN18 & 18 & 40,943&141,442&5000\\
        YAGO3-10 & 37 & 123,170 & 1,079,040 & 5000\\
        \bottomrule
        \end{tabular}
 \caption{\textbf{Data Statistics} of the benchmarks.}
 \label{tab:data_stats}
\end{table}
\section{Experiment Setup}

\paragraph{Datasets}
To evaluate our method, we conduct several experiments on four widely used KGs. To validate the accuracy of the approximations, we use smaller sized Kinship and Nations KGs for which we can make comparisons against more expensive but less approximate approaches. For the remaining experiments, we use YAGO3-10 and WN18 KGs, which are closer to real-world KGs in their size and characteristics (see Table~\ref{tab:data_stats}). 

\paragraph{Models}
We implement all methods using the same loss and optimization for training, i.e., AdaGrad and the binary cross-entropy loss. 
We use validation data to tune the hyperparameters and use a grid search to find the best hyperparameters, such as regularization parameter, and learning rate of the gradient-based method. To capture the effect of our method on link prediction task, we study the change in commonly-used metrics for evaluation in this task: mean reciprocal rank (MRR) and Hits@K. 
Further, we use the same hyperparameters as in~\citet{dettmers2017convolutional} for training link prediction models for these knowledge graphs.

\para{Influence Function}
We also compare our method with
influence function (IF)~\citep{koh2017understanding}. The influence function approximates the effect of upweighting a training sample on the loss for a specific test point. 
We use IF to approximate the change in the loss after removing a triple as: 
\begin{align}
\mathcal{I}_\text{up,loss}(\langle s', r', o \rangle, \langle s, r, o \rangle) = \phantom{\mathcal{I}_\text{up,loss}(\langle s'\rangle)}\nonumber\\ 
-\nabla_{\theta}\loss(\langle s, r, o \rangle, \hat{\theta})^{\intercal}H_{\hat{\theta}}^{-1}\nabla_{\theta}\loss(\langle s', r', o \rangle, \hat{\theta})
\end{align}
where $\langle s', r', o \rangle$ and $\langle s, r, o \rangle$ are training and test samples respectively, $\hat{\theta}$ represents the optimum parameters and $\loss(\langle s, r, o \rangle, \hat{\theta})$ represents the loss function for the test sample $\langle s, r, o \rangle$. 
Influence function does not scale well, so we only compare our method with IF on the smaller size KGs. 

\section{Experiments}
We evaluate \AALP by (\ref{sec:IF-vs-AALP})~comparing \AALP estimate with the actual effect of the attacks, (\ref{sec:exp:attack})~studying the effect of adversarial attacks on evaluation metrics, (\ref{sec:exp:intp})~exploring its application to the interpretability of KG representations, and (\ref{sec:exp:error})~detecting incorrect triples.

\begin{figure}[tb]
    \centering
        \includegraphics[width=\columnwidth]{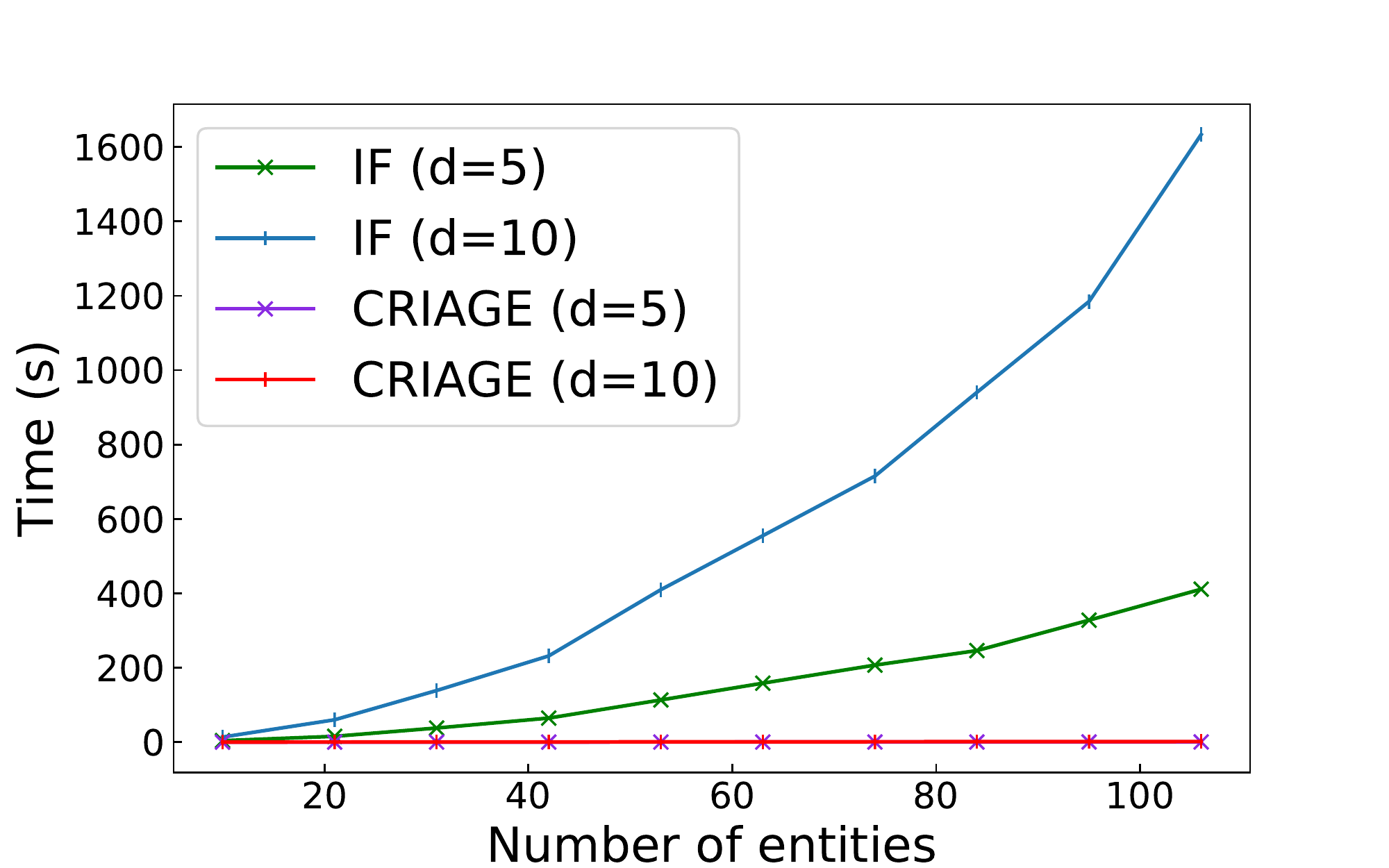}
    \caption{\textbf{Influence function vs \AALP}. We plot the average time (over $10$ facts) of influence function (IF) and \AALP to identify an adversary as the number of entities in the Kinship KG is varied (by randomly sampling subgraphs of the KG). Even with small graphs and dimensionality, IF quickly becomes impractical.} 
    \vskip -2mm
    \label{fig:in_vs_am}
\end{figure}
\begin{table*}[tb]
    \centering
    \begin{tabular}{lcccccccc}
    \toprule
        \multirow{3}{*}{\bf Methods}&\multicolumn{4}{c}{\bf             Nations}&\multicolumn{4}{c}{\bf Kinship}\\
        \cmidrule(lr){2-5}
        \cmidrule(lr){6-9}
        &\multicolumn{2}{c}{\bf             Adding}&\multicolumn{2}{c}{\bf Removing}&\multicolumn{2}{c}{\bf             Adding}&\multicolumn{2}{c}{\bf Removing}\\
        \cmidrule(lr){2-3}
        \cmidrule(lr){4-5}
        \cmidrule(lr){6-7}
        \cmidrule(lr){8-9}

         & \bf $\rho$ & \bf $\tau$ & \bf $\rho$ & \bf $\tau$  & \bf $\rho$ & \bf $\tau$ & \bf $\rho$ & \bf $\tau$  \\
    \midrule
         Ranking Based on Score &0.03&0.02&-0.01&-0.01&-0.09&-0.06&0.01&0.01\\
         Influence Function without Hessian& 0.15&0.12&0.12&0.1&0.77&0.71&0.77&0.71\\
        \AALP (Brute Force) & 0.95  & 0.84&0.94&0.85&0.99& 0.97&0.99& 0.95 \\
        Influence Function & 0.99  & 0.95&0.99  &0.96&0.99&0.98&0.99&0.98 \\
    \bottomrule
    \end{tabular}
    \caption{\textbf{Ranking modifications by their impact on the target}. We compare  the \emph{true} ranking of candidate triples with a number of approximations using ranking correlation coefficients. We compare our method with influence function (IF) with and without Hessian, and ranking the candidates based on their score, on two KGs ($d=10$, averaged over 10 random targets). For the sake of brevity, we represent the Spearman's $\rho$ and Kendall's $\tau$ rank correlation coefficients simply as $\rho$ and $\tau$.}
    \label{tab:A-m}
\end{table*}

\subsection{Influence Function vs \AALP}
\label{sec:IF-vs-AALP}
To evaluate the quality of our approximations and compare with influence function (IF), we conduct leave one out experiments. 
In this setup, we take all the neighbors of a random target triple as candidate modifications, remove them one at a time, retrain the model each time, and compute the \emph{exact} change in the score of the target triple.
We can use the magnitude of this change in score to rank the candidate triples, and compare this \emph{exact} ranking with ranking as predicted by: \AALPR{}, influence function with and without Hessian matrix, and the original model score (with the intuition that facts that the model is most confident of will have the largest impact when removed).
Similarly, we evaluate \AALPA by considering $200$ random triples that share the object entity with the target sample as candidates, and rank them as above. 

The average results of Spearman's $\rho$ and Kendall's $\tau$ rank correlation coefficients over 10 random target samples is provided in Table~\ref{tab:A-m}.
\AALP performs comparably to the influence function, confirming that our approximation is accurate. 
Influence function is slightly more accurate because they use the complete Hessian matrix over all the parameters, while we only approximate the change by calculating the Hessian over $\vo$.
The effect of this difference on scalability is dramatic, constraining IF to very small graphs and small embedding dimensionality ($d\leq 10$) before we run out of memory.
In Figure~\ref{fig:in_vs_am}, we show the time to compute a single adversary by IF compared to \AALP, as we steadily grow the number of entities (randomly chosen subgraphs), averaged over 10 random triples.
As it shows, \AALP is mostly unaffected by the number of entities while IF increases quadratically. Considering that real-world KGs have tens of thousands of times more entities, making IF unfeasible for them.


\begin{table*}[tb]
\small
  \centering
          \resizebox{\textwidth}{!}{
\begin{tabular}{clcccccccc}
\toprule
&\multirow{3}{*}{\bf Models}&\multicolumn{4}{c}{\bf YAGO3-10}&\multicolumn{4}{c}{\bf WN18}\\
\cmidrule(lr){3-6}
\cmidrule(lr){7-10}
&&\multicolumn{2}{c}{\bf All-Test}&\multicolumn{2}{c}{\bf Uncertain-Test}&\multicolumn{2}{c}{\bf All-Test}&\multicolumn{2}{c}{\bf Uncertain-Test}\\
\cmidrule(lr){3-4}
\cmidrule(lr){5-6}
\cmidrule(lr){7-8}
\cmidrule(lr){9-10}

&& \bf MRR& \bf Hits@1& \bf MRR& \bf Hits@1& \bf MRR& \bf Hits@1& \bf \bf MRR& \bf Hits@1\\ \midrule
\parbox[t]{2mm}{\multirow{6}{*}{\rotatebox[origin=c]{90}{\bf DistMult}}} &
DistMult&0.458 &37 \textcolor{red}{(0)}&1.0 &100 \textcolor{red}{(0)}&0.938 &93.1 \textcolor{red}{(0)}&1.0&100 \textcolor{red}{(0)} \\ 
& + Adding Random Attack&0.442 &34.9 \textcolor{red}{(-2.1)}&0.91&87.6 \textcolor{red}{(-12.4)}&0.926&91.1 \textcolor{red}{(-2)}&0.929&90.4 \textcolor{red}{(-9.6)}\\
& + Adding Opposite Attack&0.427 &33.2 \textcolor{red}{(-3.8)}&0.884&84.1 \textcolor{red}{(-15.9)}&0.906&87.3 \textcolor{red}{(-5.8)}&0.921&91 \textcolor{red}{(-9)}\\
\cmidrule{2-10}
& + \AALPA&0.379 &29.1 \textcolor{red}{(-7.9)}&0.71&58 \textcolor{red}{(-42)}&0.89&86.4 \textcolor{red}{(-6.7)}&0.844&81.2 \textcolor{red}{(-18.8)}\\
& + \AALP-FT&0.387 &27.7 \textcolor{red}{(-9.3)}&0.673&50.5 \textcolor{red}{(-49.5)}&0.86&79.2 \textcolor{red}{(-13.9)}&0.83&74.5 \textcolor{red}{(-25.5)}\\
& + \AALPBE&0.372 &26.9 \textcolor{red}{(-10.1)}&0.658&49.3 \textcolor{red}{(-50.7)}&0.838&77.9 \textcolor{red}{(-15.2)}&0.814&72.7 \textcolor{red}{(-27.3)}\\
\midrule
\parbox[t]{2mm}{\multirow{6}{*}{\rotatebox[origin=c]{90}{\bf ConvE}}} &
ConvE&0.497 &41.2 \textcolor{red}{(0)}&1.0 &100 \textcolor{red}{(0)}&0.94&93.3 \textcolor{red}{(0)}&1.0&100 \textcolor{red}{(0)} \\ 
& + Adding Random Attack&0.474&38.4 \textcolor{red}{(-2.8)}&0.889&83 \textcolor{red}{(-17)}&0.921 &90.1 \textcolor{red}{(-3.2)}&0.923 &89.7 \textcolor{red}{(-10.3)} \\
& + Adding Opposite Attack&0.469 &38 \textcolor{red}{(-3.2)}&0.874 &81.9 \textcolor{red}{(-18.1)}&0.915 &88.9 \textcolor{red}{(-4.4)}&0.908 &88.1 \textcolor{red}{(-11.9)} \\
\cmidrule{2-10}
& + \AALPA&0.454 &36.9 \textcolor{red}{(-4.3)}&0.738 &61.5 \textcolor{red}{(-38.5)}&0.897 &87.8 \textcolor{red}{(-5.5)}&0.895 &87.6 \textcolor{red}{(-12.4)}\\
& + \AALP-FT&0.441 &33.2 \textcolor{red}{(-8)}&0.703&57.4 \textcolor{red}{(-42.6)}&0.865&80 \textcolor{red}{(-13.3)}&0.874&79.5 \textcolor{red}{(-20.5)}\\
& + \AALPBE&0.423 & 31.9 \textcolor{red}{(-9.3)}&0.677&54.8 \textcolor{red}{(-45.2)}&0.849&79.1 \textcolor{red}{(-14.2)}&0.858&78.4 \textcolor{red}{(-21.6)}\\
\bottomrule
\end{tabular}} 
\caption{\textbf{Robustness of Representation Models}, the effect of adversarial attack on link prediction task. We consider two scenario for the target triples, 1) choosing the whole test dataset as the targets (All-Test) and 2) choosing a subset of test data that models are uncertain about them (Uncertain-Test).}
\label{tab:attack}
\end{table*}
\subsection{Robustness of Link Prediction Models}
\label{sec:exp:attack}

Now we evaluate the effectiveness of \AALP to successfully \emph{attack} link prediction by adding false facts. 
The goal here is to identify the attacks for triples in the test data, and measuring their effect on MRR and Hits@ metrics (ranking evaluations) after conducting the attack and retraining the model.

Since this is the first work on adversarial attacks for link prediction, we introduce several baselines to compare against our method. 
For finding the adversarial fact to add for the target triple $\langle s, r, o \rangle$, we consider two baselines: 1) choosing a random fake fact $\langle s', r', o \rangle$ (\textbf{Random Attack}); 2) finding $(s', r')$ by first calculating $\f(\vs, \vr)$ and then feeding $-\f(\vs, \vr)$ to the decoder of the inverter function (\textbf{Opposite Attack}). 
In addition to \AALPA, we introduce two other alternatives of our method: (1) \textbf{\AMLPFT}, that uses \AALP to \emph{increase} the score of fake fact over a test triple, i.e., we find the fake fact the model ranks second after the test triple, and identify the adversary for them, and 
(2)~\textbf{\AALPBE} that selects between \AALPA and \AMLPFT attacks based on which has a higher estimated change in score. 

\para{All-Test} 
The result of the attack on all test facts as targets is provided in the Table~\ref{tab:attack}. 
\AALPA outperforms the baselines, demonstrating its ability to effectively attack the KG representations. 
It seems DistMult is more robust against random attacks, while ConvE is more robust against designed attacks.
\AMLPFT is more effective than \AALPA since changing the score of a fake fact is easier than of actual facts; there is no existing evidence to support fake facts.
We also see that YAGO3-10 models are more robust than those for WN18. 
Looking at sample attacks (provided in Appendix~\ref{app:egs}), \AALP mostly tries to change the \emph{type} of the target object by associating it with a subject and a relation for a different entity type.

\para{Uncertain-Test} To better understand the effect of attacks, we consider a subset of test triples that 1)~the model predicts correctly, 2)~difference between their scores and the negative sample with the highest score is minimum. 
This ``Uncertain-Test'' subset contains 100 triples from each of the original test sets, and we provide results of attacks on this data in Table~\ref{tab:attack}. 
The attacks are much more effective in this scenario, causing a considerable drop in the metrics. 
Further, in addition to \AALP significantly outperforming other baselines, 
they indicate that ConvE's confidence is much more robust. 

 \begin{figure}[tb] 
  \centering
        \includegraphics[width=0.48\textwidth]{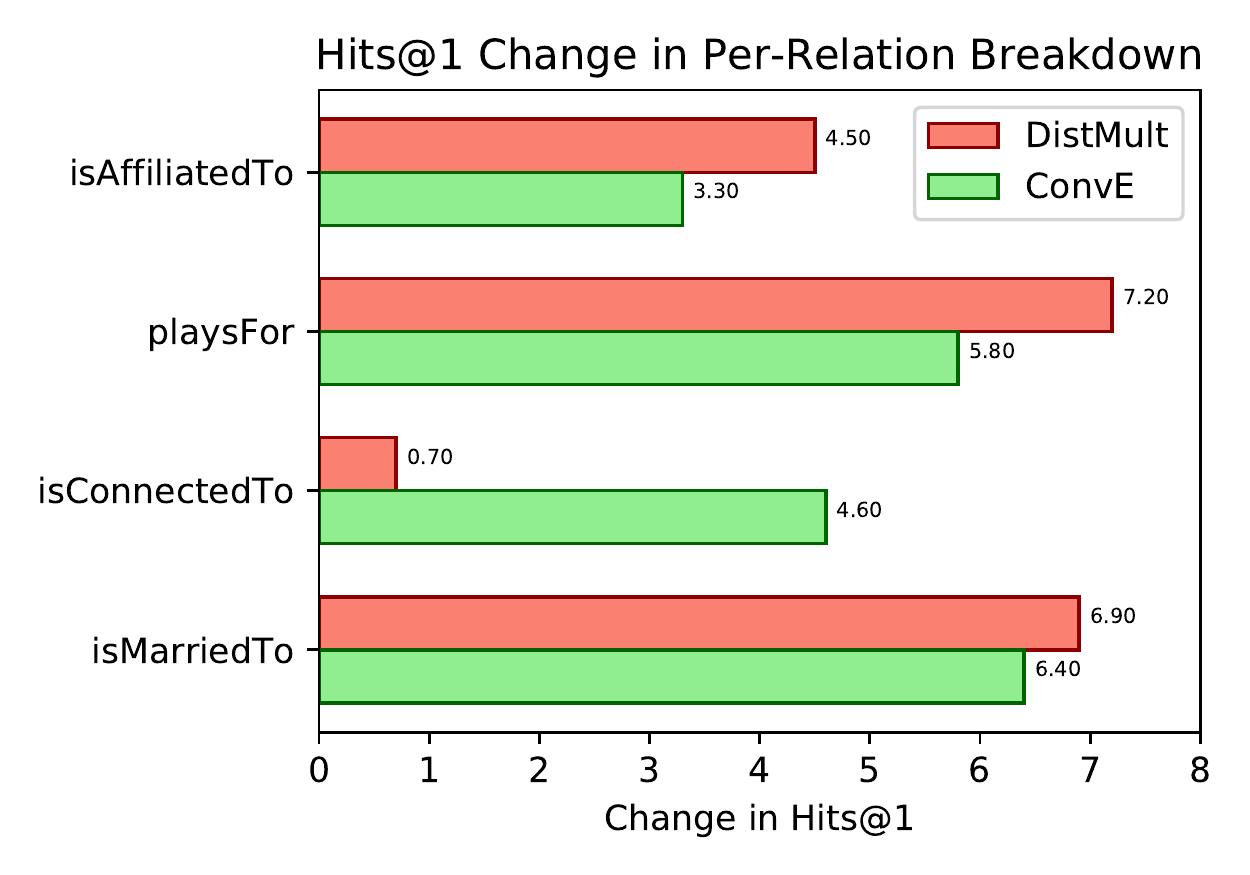}
  \caption{Per-Relation Breakdown showing the effect of \AALPA on different relations in YAGO3-10.} 
  \label{fig:per-rel}
\end{figure}

\para{Relation Breakdown}
We perform additional analysis on the YAGO3-10 dataset to gain a deeper understanding of the performance of our model. 
As shown in Figure~\ref{fig:per-rel}, both DistMult and ConvE provide a  more robust representation for \texttt{isAffiliatedTo} and \texttt{isConnectedTo} relations, demonstrating the confidence of models in identifying them. Moreover, the \AALP affects DistMult more in \texttt{playsFor} and \texttt{isMarriedTo} relations while affecting ConvE more in \texttt{isConnectedTo} relations. 

\begin{table}[tb]
    \centering
    \small
                    \setlength{\tabcolsep}{2pt}
    \begin{tabular}{ll}
    \toprule
    \multicolumn{1}{l}{\bf Rule Body, $ R_1(a,c)\wedge R_2(c,b)\Rightarrow$} &  \bf{Target}, $R(a,b)$\\
    \midrule
    {\bf Common to both} & \\
     \text{isConnectedTo}$(a,c)\wedge$ \textcolor{red!50!black}{\text{isConnectedTo}$(c,b)$} &\text{isConnectedTo} \\
     \text{isLocatedIn}$(a,c)\wedge$ \textcolor{red!50!black}{\text{isLocatedIn}$(c,b)$} &\text{isLocatedIn} \\
    \text{isAffiliatedTo}$(a,c)\wedge$ \textcolor{red!50!black}{ \text{isLocatedIn}$(c,b)$} &\text{wasBornIn}\\
    \text{isMarriedTo}$(a,c)\wedge$ \textcolor{red!50!black}{\text{hasChild}$(c,b)$}  & \text{hasChild}\\
    \addlinespace
    {\bf only in DistMult}\\
     \text{playsFor}$(a,c)\wedge$ \textcolor{red!50!black}{\text{isLocatedIn}$(c,b)$} &\text{wasBornIn}\\
     \text{dealsWith}$(a,c)\wedge$ \textcolor{red!50!black}{\text{participatedIn}$(c,b)$} &\text{participatedIn}\\
    \text{isAffiliatedTo}$(a,c)\wedge$  \textcolor{red!50!black}{\text{isLocatedIn}$(c,b)$} &\text{diedIn}\\
    \text{isLocatedIn}$(a,c)\wedge$ \textcolor{red!50!black}{\text{hasCapital}$(c,b)$}  & \text{isLocatedIn}\\
    \addlinespace
    {\bf only in ConvE} & \\
    \text{influences}$(a,c)\wedge$ \textcolor{red!50!black}{\text{influences}$(c,b)$} &\text{influences} \\
    \text{isLocatedIn}$(a,c)\wedge$ \textcolor{red!50!black}{\text{hasNeighbor}$(c,b)$} &\text{isLocatedIn} \\
    \text{hasCapital}$(a,c)\wedge$  \textcolor{red!50!black}{\text{isLocatedIn}$(c,b)$} &\text{exports}\\
    \text{hasAdvisor}$(a,c)\wedge$ \textcolor{red!50!black}{\text{graduatedFrom}$(c,b)$}  & \text{graduatedFrom}\\
    \addlinespace
    \multicolumn{2}{l}{\bf Extractions from DistMult~\citep{yang2014embedding}} \\
    \text{isLocatedIn}$(a,c)$ $\wedge$ \text{isLocatedIn}$(c,b)$ &\text{isLocatedIn}\\
    \text{isAffiliatedTo}$(a,c)$ $\wedge$ \text{isLocatedIn}$(c,b)$ &\text{wasBornIn} \\
    \text{playsFor}$(a,c)$ $\wedge$ \text{isLocatedIn}$(c,b)$  &\text{wasBornIn}\\
    \text{isAffiliatedTo}$(a,c)$ $\wedge$ \text{isLocatedIn}$(c,b)$ &\text{diedIn}\\
    \bottomrule
    \end{tabular}
    \caption{\textbf{Extracted Rules} for identifying the most influential link. We extract the patterns that appear more than $90\%$ times in the neighborhood of the target triple. The output of \AALPR is presented in \textcolor{red!50!black}{red}.}
    \label{tab:rule}
\end{table}

\subsection{Interpretability of Models}
\label{sec:exp:intp}


To be able to understand and interpret why a link is predicted using the opaque, dense embeddings, we need to find out which part of the graph was most influential on the prediction. 
To provide such explanations for each predictions, we identify the most influential fact using \AALPR{}.
Instead of focusing on individual predictions, we aggregate the explanations over the whole dataset for each relation using a simple rule extraction technique: we find simple patterns on subgraphs that surround the target triple and the removed fact from \AALPR, and appear more than $90\%$ of the time. 
We only focus on extracting length-$2$ horn rules, i.e., $R_1(a,c)\wedge R_2(c,b)\Rightarrow R(a,b)$, where $R(a,b)$ is the target and $R_2(c,b)$ is the removed fact. 

Table \ref{tab:rule} shows extracted YAGO3-10 rules that are common to both models, and ones that are not. 
The rules show several interesting inferences, such that \relLabel{hasChild} is often inferred via married parents, and \relLabel{isLocatedIn} via transitivity.
There are several differences in how the models reason as well; DistMult often uses the \relLabel{hasCapital} as an intermediate step for \relLabel{isLocatedIn}, while ConvE \emph{incorrectly} uses \relLabel{isNeighbor}.
We also compare against rules extracted by \citet{yang2014embedding} for YAGO3-10 that utilizes the structure of DistMult: they require domain knowledge on types and cannot be applied to ConvE. 
Interestingly, the extracted rules contain all the rules provided by \AALP, demonstrating that \AALP can be used to accurately interpret models, including ones that are not interpretable, such as ConvE. 
These are preliminary steps toward interpretability of link prediction models, and we leave more 
analysis of interpretability to future work.

\begin{table}[]
    \centering
    \small
    \begin{tabular}{lcccc}
    \toprule
        \multirow{2}{*}{\bf Methods}&\multicolumn{2}{c}{\bf             $\langle s', r', o \rangle$ Noise}&\multicolumn{2}{c}{\bf $\langle s', r, o \rangle$ Noise}\\
        \cmidrule(lr){2-3}
        \cmidrule(lr){4-5}

         &  Hits@1 &  Hits@2 &  Hits@1 &  Hits@2 \\
    \midrule
             Random&19.7&39.4&19.7&39.4\\
         Lowest& 16&37&26&47\\
        \AALP &  42&62&55&76\\
    \bottomrule
    \end{tabular}
    \caption{\textbf{Error Detection Accuracy} in the neighborhood of 100 chosen samples. We choose the neighbor with the least value of $\Delta_{(s',r')}(s,r,o)$ as the incorrect fact. This experiment assumes we know each target fact has exactly one error. }
    \label{tab:e-d}
\end{table}

\subsection{Finding Errors in Knowledge Graphs}
\label{sec:exp:error}

Here, we demonstrate another potential use of adversarial modifications: finding erroneous triples in the knowledge graph. 
Intuitively, if there is an error in the graph, the triple is likely to be inconsistent with its neighborhood, and thus the model should put least trust on this triple. 
In other words, the error triple should have the least influence on the model's prediction of the training data.
Formally, to find the incorrect triple $\langle s', r', o\rangle$ in the neighborhood of the train triple $\langle s, r, o\rangle$, we need to find the triple $\langle s',r',o\rangle$ that results in the \emph{least} change $\Delta_{(s',r')}(s,r,o)$ when removed from the graph.

To evaluate this application, we inject random triples into the graph, and measure the ability of \AALP to detect the errors using our optimization.
We consider two types of incorrect triples: 1) incorrect triples in the form of $\langle s', r, o\rangle$ where $s'$ is chosen randomly from all of the entities, and 2) incorrect triples in the form of $\langle s', r', o\rangle$ where $s'$ and $r'$ are chosen randomly. 
We choose $100$ random triples from the observed graph, and for each of them, add an incorrect triple (in each of the two scenarios) to its neighborhood. 
Then, after retraining DistMult on this noisy training data, we identify error triples through a search over the neighbors of the $100$ facts. 
The result of choosing the neighbor with the least influence on the target is provided in the Table~\ref{tab:e-d}. 
When compared with baselines that randomly choose one of the neighbors, or assume that the fact with the lowest score is incorrect, 
we see that \AALP outperforms both of these with a considerable gap, obtaining an accuracy of $42\%$ and $55\%$ in detecting errors. 
\section{Related Work}
Learning relational knowledge representations has been a focus of active research in the past few years, but to the best of our knowledge, this is the first work on conducting adversarial modifications on the link prediction task. 

\para{Knowledge graph embedding}
 There is a rich literature on representing knowledge graphs in vector spaces that differ in their scoring functions~\citep{wang2017knowledge, goyal2018graph, fooshee2018deep}. Although \AALP is primarily applicable to multiplicative scoring functions~\citep{nickel2011three, socher2013reasoning, yang2014embedding,trouillon2016complex}, these ideas apply to additive scoring functions~\citep{bordes13:translating,wang2014knowledge,lin2015learning,nguyen2016stranse} as well, as we show in Appendix~\ref{app_transe}.

Furthermore, there is a growing body of literature that incorporates an extra types of evidence for more informed embeddings 
such as numerical values~\citep{garcia2017kblrn}, images~\citep{onoro2017representation}, text~\citep{toutanova2015representing,toutanova2016compositional,tu2017cane}, and their combinations~\citep{pezeshkpour2018embedding}.
Using \AALP, we can gain a deeper understanding of these methods, especially those that build their embeddings wit hmultiplicative scoring functions.

\para{Interpretability and Adversarial Modification}
There has been a significant recent interest in conducting an adversarial attacks on different machine learning models \citep{biggio2014security,papernot2016limitations,dong2017towards,zhao2018data,zhengli2018iclr,brunet2018understanding} to attain the interpretability, and further, evaluate the robustness of those models. 
\citet{koh2017understanding} uses influence function to provide an approach to understanding black-box models by studying the changes in the loss occurring as a result of changes in the training data. 
In addition to incorporating their established method on KGs, we derive a novel approach that differs from their procedure in two ways: 
(1) instead of changes in the loss, we consider the changes in the scoring function, which is more appropriate for KG representations, and 
(2) in addition to searching for an attack, we introduce a gradient-based method that is much faster, especially for ``adding an attack triple'' (the size of search space make the influence function method infeasible). 
Previous work has also considered adversaries for KGs, but as part of training to improve their representation of the graph~\citep{minervini2017adversarial,cai2018kbgan}.  

\para{Adversarial Attack on KG} 
Although this is the first work on adversarial attacks for link prediction, 
there are two approaches \citep{dai2018adversarial, zugner2018adversarial} that consider the task of adversarial attack on graphs. 
There are a few fundamental differences from our work: 
(1) they build their method on top of a path-based representations while we focus on embeddings, 
(2) they consider node classification as the target of their attacks while we attack link prediction, and 
(3) they conduct the attack on small graphs due to restricted scalability, while the complexity of our method does not depend on the size of the graph, but only the neighborhood, allowing us to attack real-world graphs.  

\section{Conclusions}
Motivated by the need to analyze the robustness and interpretability of link prediction models, we present a novel approach for conducting adversarial modifications to knowledge graphs. 
We introduce \AALP, completion robustness and interpretability via adversarial graph edits: identifying the fact to add into or remove from the KG that changes the prediction for a target fact. 
\AALP uses (1)~an estimate of the score change for any target triple after adding or removing another fact, and (2)~a gradient-based algorithm for identifying the most influential modification. 
We show that \AALP can effectively reduce ranking metrics on link prediction models upon applying the attack triples. Further, we incorporate the \AALP to study the interpretability of KG representations by summarizing the most influential facts for each relation. 
Finally, using \AALP, we introduce a novel automated error detection method for knowledge graphs.
We have release the open-source implementation of our models at: \url{https://pouyapez.github.io/criage}. 

\section*{Acknowledgements}
We would like to thank Matt Gardner, Marco Tulio Ribeiro, Zhengli Zhao, Robert L. Logan IV, Dheeru Dua and the anonymous reviewers for their detailed feedback and suggestions.  This work is supported in part by Allen Institute for Artificial Intelligence (AI2) and in part by NSF awards \#IIS-1817183 and \#IIS-1756023. The views expressed are those of the authors and do not reflect the official policy or position of the funding agencies.

\bibliography{main}
\bibliographystyle{plainnat}

\appendix
\section{Appendix}
We approximate the change on the score of the target triple upon applying attacks other than the $\langle s', r', o \rangle$ ones. Since each relation appears many times in the training triples, we can assume that applying a single attack will not considerably affect the relations embeddings. As a result, we just need to study the attacks in the form of $\langle s, r', o \rangle$ and $\langle s, r', o' \rangle$. 
Defining the scoring function as $\psi(s,r,o) = \f(\vs, \vr) \cdot \vo= \z_{s,r} \cdot \vo$, we further assume that $\psi(s,r,o) =\vs \cdot \g(\vr, \vo) =\vs \cdot \x_{r,o}$.

\begin{table*}[tb]
    \centering
    \small
    \begin{tabular}{lcc}
    \toprule
   & \multicolumn{1}{c}{\bf Target Triple} &  \bf \AALPA \\
    \midrule
     \parbox[t]{2mm}{\multirow{4}{*}{\rotatebox[origin=c]{90}{\bf DistMult}}}
    &Brisbane Airport, isConnectedTo, Boulia Airport&Osman Ozköylü, isPoliticianOf, Boulia Airport\\
    &Jalna District, isLocatedIn, India&United States, hasWonPrize, India\\
    &Quincy Promes, wasBornIn, Amsterdam&Gmina Krzeszyce, hasGender, Amsterdam\\
    &Princess Henriette, hasChild, Violante Bavaria&Al Jazira Club, playsFor, Violante Bavaria\\
    \midrule
 \parbox[t]{2mm}{\multirow{4}{*}{\rotatebox[origin=c]{90}{\bf ConvE}}} 
     &Brisbane Airport, isConnectedTo, Boulia Airport&Victoria Wood, wasBornIn, Boulia Airport\\
    &National Union(Israel), isLocatedIn, Jerusalem&Sejad Halilović, isAffiliatedTo, Jerusalem\\
     &Robert Louis, influences, David Leavitt&David Louhoungou, hasGender, David Leavitt\\
    &Princess Henriette, hasChild, Violante Bavaria&Jonava, isAffiliatedTo, Violante Bavaria\\

    \bottomrule
    \end{tabular}
    \caption{Top adversarial triples for target samples.}
    \label{tab:int}
\end{table*}

\subsection{Modifications of the Form $\langle s, r', o' \rangle $}
\label{app:srpop}
Using similar argument as the attacks in the form of $\langle s', r', o \rangle$, we can calculate the effect of the attack, 
$\overline{\psi}{(s,r,o)}-\psi(s, r, o)$ as: 
\begin{align}
\overline{\psi}{(s,r,o)}-\psi(s, r, o)=(\vsn-\vs) \x_{s, r}
\end{align}
where $\x_{s, r} = \g(\vr,\vo)$. 

We now derive an efficient computation for $(\vsn-\vs)$. 
First, the derivative of the loss $\loss(\overline{G})= \loss(G)+\loss(\langle s, r', o' \rangle)$ over $\vs$ is:
 \begin{align}
    \nabla_{e_s} \loss(\overline{G}) = \nabla_{e_s} \loss(G) - (1-\varphi) \x_{r', o'} 
 \end{align}
 where $\x_{r', o'} = \g(\vr',\vo')$, and $\varphi = \sigma(\psi(s,r',o'))$. 
At convergence, after retraining, we expect $\nabla_{e_s} \loss(\overline{G})=0$.
We perform first order Taylor approximation of $\nabla_{e_s} \loss(\overline{G})$ to get:
\begin{align}
    0 \simeq &- (1-\varphi)\x_{r',o'}^\intercal+&\nonumber\\
    &(H_s+\varphi(1-\varphi)\x_{r',o'}^{\intercal} \x_{r',o'})(\vsn-\vs)&
\end{align}
where $H_s$ is the $d\times d$ Hessian matrix for $s$, i.e. second order derivative of the loss w.r.t. $\vs$, computed sparsely.
Solving for $\vsn-\vs$ gives us:
\begin{align}
    &\vsn-\vs = \nonumber\\
    &(1-\varphi) (H_s + \varphi (1-\varphi) \x_{r',o'}^\intercal\x_{r',o'})^{-1} \x_{r',o'}^\intercal \nonumber
\end{align}
In practice, $H_s$ is positive definite, making $H_s + \varphi (1-\varphi) \x_{r',o'}^\intercal\x_{r',o'}$ positive definite as well, and invertible. Then, we compute the score change as: 
\begin{align}
       &\overline{\psi}{(s,r,o)}-\psi(s, r, o)= \x_{r,o} (\vsn-\vs) = &\nonumber\\
       & ((1-\varphi) (H_s + \varphi (1-\varphi) \x_{r',o'}^\intercal \x_{r',o'})^{-1} \x_{r',o'}^\intercal )\x_{r,o}.&
\end{align}

\subsection{Modifications of the Form $\langle s, r', o \rangle $}
\label{app:srpo}

In this section we approximate the effect of attack in the form of $\langle s, r', o \rangle $. In contrast to $\langle s', r', o \rangle$ attacks, for this scenario we need to consider the change in the $\vs$, upon applying the attack, in approximation of the change in the score as well. Using previous results, we can approximate the $\von-\vo$ as: 
\begin{align}
    \von-\vo &=\nonumber\\ &(1-\varphi) (H_o + \varphi (1-\varphi) \z_{s,r'}^\intercal \z_{s,r'})^{-1} \z_{s,r'}^\intercal 
\end{align}
and similarly, we can approximate $\vsn-\vs$ as: 
\begin{align}
    \vsn-\vs &=\nonumber\\ & (1-\varphi) (H_s + \varphi (1-\varphi) \x_{r',o}^\intercal \x_{r',o})^{-1} \x_{r',o}^\intercal 
\end{align}
where $H_s$ is the Hessian matrix over $\vs$. Then using these approximations:
\begin{align}
 &\z_{s,r}(\von-\vo) =\nonumber\\&  \z_{s,r} ((1-\varphi) (H_o + \varphi (1-\varphi) \z_{s,r'}^\intercal \z_{s,r'})^{-1}  \z_{s,r'}^\intercal )\nonumber
 \end{align}
and:
\begin{align}
  &(\vsn-\vs) \x_{r,\Bar{o}}=\nonumber\\& ((1-\varphi) (H_s + \varphi (1-\varphi) \x_{r',o}^\intercal \x_{r',o})^{-1} \x_{r',o}^\intercal ) \x_{r,\Bar{o}}\nonumber
\end{align}
and then calculate the change in the score as: 
\begin{align}
       &\overline{\psi}{(s,r,o)}-\psi(s, r, o)= \nonumber\\&  \z_{s,r}.(\von-\vo) +(\vsn-\vs).\x_{r,\Bar{o}} 
       =  \nonumber\\& \z_{s,r} ((1-\varphi) (H_o + \varphi (1-\varphi) \z_{s,r'}^\intercal \z_{s,r'})^{-1} \z_{s,r'}^\intercal )+ \nonumber\\&
        ((1-\varphi) (H_s + \varphi (1-\varphi) \x_{r',o}^\intercal \x_{r',o})^{-1} \x_{r',o}^\intercal ) \x_{r, \Bar{o}}
\end{align}

\subsection{First-order Approximation of the Change For TransE}
\label{app_transe}

In here we derive the approximation of the change in the score upon applying an adversarial modification for TransE~\citep{bordes13:translating}. Using similar assumptions and parameters as before, to calculate the effect of the attack, 
$\overline{\psi}{(s,r,o)}$ (where $\psi{(s,r,o)}=|\vs+\vr-\vo|$), we need to compute $\von$. To do so, we need to derive an efficient computation for $\von$. 
First, the derivative of the loss $\loss(\overline{G})= \loss(G)+\loss(\langle s', r', o \rangle)$ over $\vo$ is:
 \begin{align}
    \nabla_{e_o} \loss(\overline{G}) = \nabla_{e_o} \loss(G) + (1-\varphi) \frac{\z_{s', r'}-\vo}{\psi(s',r',o)} 
 \end{align}
 where $\z_{s', r'} = \vs'+ \vr'$, and $\varphi = \sigma(\psi(s',r',o))$. 
At convergence, after retraining, we expect $\nabla_{e_o} \loss(\overline{G})=0$.
We perform first order Taylor approximation of $\nabla_{e_o} \loss(\overline{G})$ to get:
\begin{align}
    &0 \simeq &\nonumber\\&
    (1-\varphi) \frac{(\z_{s', r'}-\vo)^\intercal}{\psi(s',r',o)}+(H_o - H_{s',r',o})(\von-\vo)&\\\nonumber&
    H_{s',r',o} = (1-\varphi)\varphi \frac{(\z_{s', r'}-\vo)^\intercal(\z_{s', r'}-\vo)}{\psi(s',r',o)^2}+ &\\& \frac{1-\varphi}{\psi(s',r',o)}-(1-\varphi) \frac{(\z_{s', r'}-\vo)^\intercal(\z_{s', r'}-\vo)}{\psi(s',r',o)^3}&
\end{align}
where $H_o$ is the $d\times d$ Hessian matrix for $o$, i.e., second order derivative of the loss w.r.t. $\vo$, computed sparsely.
Solving for $\von$ gives us:
\begin{align}
    &\von = -(1-\varphi) (H_o - H_{s',r',o})^{-1} \frac{(\z_{s', r'}-\vo)^\intercal}{\psi(s',r',o)}\nonumber\\& + \vo
\end{align}
Then, we compute the score change as: 
\begin{align}
       &\overline{\psi}{(s,r,o)}= |\vs+\vr-\von|\nonumber\\
       &= |\vs+\vr+(1-\varphi) (H_o - H_{s',r',o})^{-1} \nonumber\\& \frac{(\z_{s', r'}-\vo)^\intercal}{\psi(s',r',o)} - \vo|
\end{align}

Calculating this expression is efficient since $H_o$ is a $d\times d$ matrix.

\subsection{Sample Adversarial Attacks}
\label{app:egs}
In this section, we provide the output of the \AALPA for some target triples. Sample adversarial attacks are provided in Table~\ref{tab:int}. As it shows, \AALPA attacks mostly try to change the \emph{type} of the target triple's object by associating it with a subject and a relation that require a different entity types.

\end{document}